\documentclass[10pt, conference, compsocconf]{IEEEtran}
\IEEEoverridecommandlockouts
\ifCLASSINFOpdf
\else
\fi
\usepackage[numbers]{natbib}
\usepackage{flushend}
\usepackage{fixltx2e}
\usepackage{mathtools}
\usepackage{balance}
\usepackage{ctable}

\usepackage{bbm}



\makeatletter
\newcommand*\titleheader[1]{\gdef\@titleheader{#1}}
\AtBeginDocument{%
	\let\st@red@title\@title%
	\def\@title{%
		\bgroup\normalfont\large\centering\@titleheader\par\egroup
		\vskip1.5em\st@red@title}
}
\makeatother

\title{Mined Semantic Analysis: A New Concept Space Model \\for Semantic Representation of Textual Data\thanks{This work was partially supported by the National Science Foundation under grant number 1624035. Any opinions, findings, and conclusions or recommendations expressed in this material are those of the author(s) and do not necessarily reflect the views of the National Science Foundation.}}

\titleheader{Accepted at 2017 IEEE International Conference on Big Data (BIGDATA)}

\begin{document}


\author{\IEEEauthorblockN{Walid Shalaby, Wlodek Zadrozny}
\IEEEauthorblockA{Department of Computer Science\\
University of North Carolina at Charlotte\\
Charlotte, USA\\
\{wshalaby, wzadrozn\}@uncc.edu}}


%


\IEEEoverridecommandlockouts
\IEEEpubid{\makebox[\columnwidth]{xxx-x-xxxx-xxxx-x/17/\$31.00~
		\copyright2017
		IEEE \hfill} \hspace{\columnsep}\makebox[\columnwidth]{ }}

\maketitle

\begin{abstract}
Mined Semantic Analysis ({\it MSA}) is a novel concept space model which employs unsupervised learning to generate semantic representations of text. {\it MSA} represents textual structures (terms, phrases, documents) as a Bag of Concepts (BoC) where concepts are derived from concept rich encyclopedic corpora. Traditional concept space models exploit only target corpus content to construct the concept space. {\it MSA}, alternatively, uncovers implicit relations between concepts by mining for their associations (e.g., mining Wikipedia's "See also" link graph). We evaluate {\it MSA}'s performance on benchmark datasets for measuring semantic relatedness of words and sentences. Empirical results show competitive performance of {\it MSA} compared to prior state-of-the-art methods. Additionally, we introduce the first analytical study to examine statistical significance of results reported by different semantic relatedness methods. Our study shows that, the nuances of results across top performing methods could be statistically insignificant. The study positions {\it MSA} as one of state-of-the-art methods for measuring semantic relatedness, besides the inherent interpretability and simplicity of the generated semantic representation.

\end{abstract}


%
\IEEEpeerreviewmaketitle

\section{Introduction}

\begin{figure*}
	\centering
	\includegraphics[width=9.38cm,height=9.38cm,keepaspectratio]{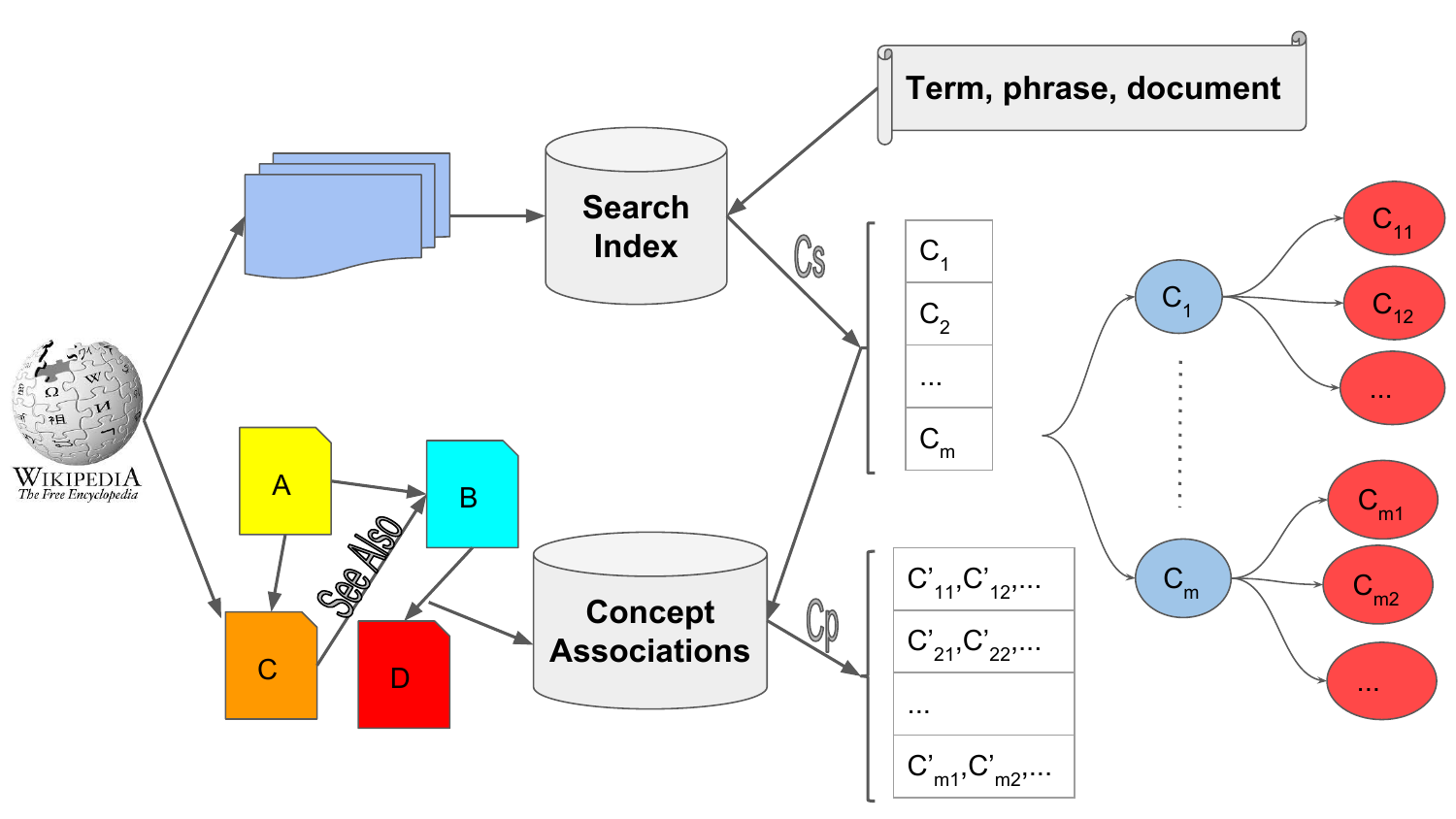}
	\caption{{\it MSA} generates the concept vector of a given textual structure through: 1) explicit concept retrieval from the index (top), and 2) concept expansion from the concept-concept associations repository (bottom).}
	\label{arch}
\end{figure*} 

Semantic representation of textual content has gained enormous attention within the Natural Language Processing (NLP) community as a means for automating Natural Language Understanding (NLU). To this end, several semantic representation paradigms have evolved over decades of research in NLP in order to overcome some of the inherent linguistic and computational limitations of the traditional Bag of Words (BoW) model. 

One of the most prominent research areas in NLU is the distributional semantics models. These models are inspired by the distributional hypothesis \cite{harris1954distributional}  which emphasizes the idea that similar words tend to appear in similar contexts and thus have similar contextual distributions. Therefore the meaning of words can be determined by analyzing the statistical patterns of word usage in various contexts.


Another active line of research is concerned with explicit semantic representations of text as a Bag of Concepts (BoC) through text conceptualization. Such methods focus on the global contexts of terms (i.e., documents in which they appeared), or their properties in existing KBs in order to figure out their meanings. Text conceptualization is motivated by the idea that humans understand languages through multi-step cognitive processes which involves building rich models of the world and making multilevel generalizations from the input text \cite{understanding-short-texts}. One way of automating such generalizations is through text conceptualization; either by extracting basic level concepts from the input text using concept KBs \cite{kim2013context,song2015open}, or mapping the whole input into a concept space that captures its semantics as in ESA \cite{gabrilovich2007computing}. 

Measuring the semantic similarity/relatedness between text structures (words, phrases, and documents) has been the standard evaluation task for almost all proposed semantic representation models. 
Although semantic similarity and relatedness are often used interchangeably in the literature, they do not represent the same task \cite{budanitsky2006evaluating}. Evaluating genuine similarity is, and should be, concerned with measuring the similarity or resemblance in meanings and hence focuses on the synonymy relations (e.g., \textit{smart},\textit{intelligent}). Relatedness, on the other hand, is more general and covers broader scope as it focuses on other relations such as antonymy (\textit{old},\textit{new}), hypernymy (\textit{red},\textit{color}), and other functional associations (\textit{money},\textit{bank}).

Semantic relatedness has many applications in NLP and Information Retrieval (IR) for addressing problems such as word sense disambiguation, paraphrasing, text categorization, semantic search, and others. 


In this paper we propose Mined Semantic Analysis (\textit{MSA}), a novel concept space model for semantic representation of textual data using unsupervised data mining techniques. {\it MSA} represents textual structures (terms, phrases, documents) as a BoC. Unlike other  concept space models which look for direct associations between concepts and terms through statistical co-occurrence \cite{camacho2015nasari,hassan2011semantic,gabrilovich2007computing}, \textit{MSA} discovers implicit concept-concept associations using rule mining \cite{agrawal1994fast}. \textit{MSA} uses these associations subsequently to enrich the terms concept vector with these associated concepts. 

\textit{MSA} utilizes a search index created using concept rich corpora (e.g., {\it Wikipedia}). The concept vector of a given text is constructed through two phases. First, an initial set of candidate concepts is retrieved from the index. Second, the candidates set is augmented with other related concepts from the discovered concept-concept association rules. Following this strategy, \textit{MSA} identifies not only concepts directly related to a given text but also other concepts associated implicitly with them. 

The contributions of this paper are threefold: First, we introduce a novel concept space model for semantic representation which augments explicit semantics with conceptual associations through data mining techniques. Second, we demonstrate the effectiveness of this method for evaluating semantic relatedness on various benchmark datasets. Third, we present the first analytical study to examine statistical significance of results reported by different semantic relatedness methods. Our study shows that, 
the nuances of results across top performing methods could be statistically insignificant. The study positions \textit{MSA} as one of state-of-the-art methods for measuring semantic relatedness.

\section{Related Work}
Several vector-based semantic representation models have been proposed in the literature. Those models represent textual structures as "meaning" vectors. As pointed out by \citet{baroni2014don}, those vectors are either estimated by means of statistical modeling such as \textit{LSA} \cite{landauer1997well} and \textit{LDA} \cite{blei2003latent}, or more recently through neural network based representations such as \textit{CW} \cite{collobert2008unified}, \textit{Word2Vec} \cite{mikolov2013efficient}, and \textit{GloVe} \cite{pennington2014glove}. 

Knowledge-based models were also proposed for measuring semantic relatedness \cite{budanitsky2006evaluating,pilehvar2015senses}. Those models utilize dictionaries such as Wordnet \cite{wordnet} and Wiktionary\footnote{https://www.wiktionary.org}, and use explicit word relations to infer semantic relatedness. 

Explicit concept space models such as \textit{ESA} \cite{gabrilovich2007computing}, \textit{SSA} \cite{hassan2011semantic}, and \textit{NASARI} \cite{camacho2015nasari} construct Bag of Concepts (BoC) vectors to represent textual structures using concepts in encyclopedic knowledge bases such as \textit{Wikipedia}. Those {\it BoC} vectors capture the main topics of the given text and therefore are useful for understanding its semantics. 

The BoC representations have proven efficacy for semantic analysis of textual data especially short texts where contextual information is missing or insufficient. For example, measuring lexical semantic similarity/relatedness \cite{gabrilovich2007computing, li2013computing,kim2013context}, text categorization \cite{song2014dataless}, search and relevancy ranking \cite{egozi2011concept}, search query understanding \cite{wang2015query}, short text segmentation \cite{hua2015short}, and others. Semantic relatedness models typically employ those semantic vectors to measure relatedness using appropriate similarity measure between the vectors.  

A closely related method to {\it MSA} is Explicit Semantic Analysis (\textit{ESA}) \cite{gabrilovich2007computing}. \textit{ESA} constructs the concept space of a term by searching an inverted index of term-concept co-occurrences. \textit{ESA} is mostly the traditional vector space model applied to \textit{Wikipedia} articles. 
\textit{ESA} is effective in retrieving concepts which explicitly mention the target search terms in their content. However, it fails to identify other implicit concepts which do not contain the search terms. 
\textit{MSA} bridges this gap by mining for concept-concept associations and thus augmenting the concept space identified by \textit{ESA} with more relevant concepts.

Salient Semantic Analysis (\textit{SSA}) was proposed by \citet{hassan2011semantic} and uses \textit{Wikipedia} concepts to build semantic profiles of words. \textit{SSA} is more conservative than \textit{ESA} as it defines word meaning by its immediate context and therefore might yield concepts of higher relevancy. However, it is still limited to surface semantic analysis because it, like \textit{ESA}, utilizes only direct associations between words and concepts and fails to capture other implicit concepts not directly co-occurring with target words in the same context.

Another closely related model is Latent Semantic Analysis (\textit{LSA}) \cite{deerwester1990indexing,landauer1997well}. \textit{LSA} is a statistical model that was originally proposed to solve the vocabulary mismatch problem in IR. \textit{LSA} first builds a term-document co-occurrence matrix from textual corpus and then maps that matrix into a new space using singular-value decomposition. In that semantic space terms and documents that have similar meaning will be placed close to one another. Though its effectiveness, \textit{LSA} has been known to be hard to explain because it is difficult to map the computed space dimensions into meaningful concepts. \textit{MSA}, alternatively, generates explicit conceptual mappings that are interpretable by humans making it more intuitive than \textit{LSA}.

\begin{table}
	\centering
	\small
	\caption{The concept representation of "Computational Linguistics"}	
	\begin{tabular}{l|l}
		\specialrule{.1em}{.05em}{0.05em}
		\textbf{Explicit Concepts}             & \textbf{Implicit Concepts}             \\ \hline
		Parse tree               & Universal Networking Language   \\ 
		Temporal annotation      & Translation memory              \\ 
		Morphological dictionary & Systemic functional linguistics \\ 
		Textalytics              & Semantic relatedness            \\ 
		Bracketing               & Quantitative linguistics        \\ 
		Lemmatization            & Natural language processing     \\ 
		Indigenous Tweets        & Internet linguistics            \\ 
		Statistical semantics    & Grammar induction               \\ 
		Treebank                 & Dialog systems                  \\ 
		Light verb               & Computational semiotics         \\ 
		\specialrule{.1em}{.1em}{0.1em}
	\end{tabular}
	\label{cl}
\end{table}

\section{Mined Semantic Analysis}
We call our approach Mined Semantic Analysis (\textit{MSA}) as it utilizes data mining techniques in order to discover the concept space of textual structures. The motivation behind our approach is to mitigate a notable gap in previous concept space models which are limited to direct associations between words and concepts. Therefore those models lack the ability to transfer the association relation to other implicit concepts which contribute to the meaning of these words.

Figure \ref{arch} shows {\it MSA}'s architecture. In a nutshell, {\it MSA} generates the concept vector of a given text by utilizing two repositories created offline: 1) a search index of {\it Wikipedia} articles, and 2) a concept-concept associations repository created by mining the \textit{"See also"} link graph of \textit{Wikipedia}. First, the explicit concept space is constructed by retrieving concepts (titles of articles) explicitly mentioning the given text. Second, implicit concepts associated with each of the explicit concepts are retrieved from the associations repository and used to augment the concept space.

To demonstrate our approach, we provide an example of exploring the concept space of \textit{"Computational Linguistics"} (Table \ref{cl}). Column 1 shows the explicit concepts retrieved by searching \textit{Wikipedia}\footnote{We search \textit{Wikipedia} using a term-concept inverted index and limit the search space to articles with min. length of 2k and max. title length of 3 words.}. Column 2 shows the same explicit concepts in column 1 enriched by implicit concepts. As we can notice, those implicit concepts could augment the explicit concept space by more related concepts which contribute to understanding \textit{"Computational Linguistics"}. It is worth mentioning that not all implicit concepts are equally relevant, therefore we also propose an automated mechanism for ranking those concepts in a way that reflects their relatedness to the original search term.

\subsection{The Search Index}
\textit{MSA} starts constructing the concept space of term(s) by searching for an initial set of candidate explicit concepts. For this purpose, we build a search index of a concept rich corpus such as {\it Wikipedia} where each article represents a concept. This is similar to the idea of the inverted index introduced in \textit{ESA} \cite{gabrilovich2007computing}. We build the index using \textit{Apache Lucene}\footnote{http://lucene.apache.org/core/}, an open-source indexing and search engine. For each article we index the title, content, length, \# of outgoing links, and the \textit{"See also"} section.

During search we use some parameters to tune the search space. Specifically, we define the following parameters to provide more control over search: 

\textit{\textbf{Article Length ($L$)}}: Minimum length of the article. 


\noindent\textit{\textbf{Outdegree ($O$)}}: Minimum \# of outgoing links per article.

\noindent\textit{\textbf{Title Length ($\tau$)}}: Used prune articles that have long titles. It represents the maximum \# of words in the title. 

\noindent\textit{\textbf{Number of Concepts ($M$)}}: Maximum \# of concepts (articles) to retrieve as initial candidate concepts.

\subsection{Association Rules Mining}
In order to discover the implicit concepts, we employ the Apriori algorithm for association rule learning \cite{agrawal1994fast} to learn implicit relations between concepts using {\it Wikipedia}'s \textit{"See also"} link graph. 

Formally, given a set of concepts \begin{math} C = \{c_1,c_2,...,c_N\}\end{math} of size \textit{N} (i.e., all \textit{Wikipedia} articles). We build a dictionary of transactions \begin{math}T = \{t_1,t_2,t_3,...,t_M\}\end{math} of size \textit{M} such that $M\!\le\!N$. Each transaction \textit{t} $\in$ \textit{T} contains a subset of concepts in \textit{C}. \textit{t} is constructed from each article in \textit{Wikipedia} that contains at least one entry in its \textit{"See also"} section. For example, if an article representing concept \begin{math}c_1\end{math} with entries in its \textit{"See also"} section referring to concepts \begin{math}\{c_2,c_3,...,c_n\}\end{math}, a transaction \begin{math}t\!=\!\{c_1,c_2,c_3,...,c_n\}\end{math} will be learned and added to \textit{T}. A set of rules \textit{R} is then created by mining \textit{T}. Each rule \textit{r} $\in$ \textit{R} is defined as in equation 1:
\begin{equation}
r(s,f)=\{(X\Rightarrow{Y}) : X,Y\!\subseteq\!C \ and\ X\!\cap\!Y\!=\!\emptyset\}
\end{equation}
Both \textit{X} and \textit{Y} are subsets of concepts in \textit{C}. \textit{X} are called the antecedents of \textit{r} and \textit{Y} are called the consequences. Rule \textit{r} is parameterized by two parameters: 1) \emph{support} (\textit{s}) which indicates how many times both \textit{X} and \textit{Y} co-occur in \textit{T}, and 2) \emph{confidence} (\textit{f}) which is \textit{s} divided by number of times \textit{X} appeared in \textit{T}.

After learning \textit{R}, we end up having concept(s)-concept(s) associations. Using such rules, we can determine the strength of those associations based on {\it s} and {\it f}. 

As the number of rules grows exponentially with the number of concepts, we define the following parameters to provide more fine grained control on participating rules during explicit concept expansion:

\noindent\textit{\textbf{Consequences Size ($|Y|$)}}: \# of concepts in rule consequences. 

\noindent\textit{\textbf{Support Count ($\sigma$)}}: Defines the minimum \# of times antecedent concept(s) should appear in \textit{T}.

\noindent\textit{\textbf{Minimum Support ($\epsilon$)}}: Defines the minimum strength of the association between rule concepts. For example, if $\epsilon\!=\!2$, then all rules whose support $s>=2$ will be considered during concept expansion.

\noindent\textit{\textbf{Minimum Confidence ($\upsilon$)}}: Defines the minimum strength of the association between rule concepts compared to other rules with same antecedents. For example, if $\upsilon\!=\!0.5$, then all rules whose confidence $f>=0.5$ will be considered during concept expansion. 

\subsection{Constructing the Concept Space}
Given a set of concepts \textit{C} of size \textit{N}, \textit{MSA} constructs the bag-of-concepts vector \textit{C\textsubscript{t}} of term(s) \textit{t} through two phases: {\it Search} and {\it Expansion}. In the search phase, \textit{t} is represented as a search query and is searched for in the \textit{Wikipedia} search index. 
This returns a weighted set of articles that best matches \textit{t} based on the vector space model. We call the set of concepts representing those articles \textit{C\textsubscript{s}} and is represented as in equation 2:
\begin{equation}
\begin{multlined}
C_s = \{(c_i,w_i) : c_i \in C\ and\ i<=N\} \\
subject \ to: |title(c_i)|<=\tau,\
length(c_i)>=L,\\ O_{c_i}<=O,\ |C_s|<=M
\end{multlined}
\end{equation}
Note that we search all articles whose content length, title length, and outdegree meet the thresholds \textit{L}, \textit{$\tau$}, \textit{$O$} respectively. The weight of \textit{c\textsubscript{i}} is denoted by \textit{w\textsubscript{i}} and represents the match score between \textit{t} and \textit{c\textsubscript{i}} as returned by Lucene.

In the expansion phase, we first prune all the search concepts whose support count is below the threshold $\sigma$. We then use the learned association rules to expand each remaining concept \textit{c} in \textit{C\textsubscript{s}} by looking for its associated set of concepts in \textit{R}. Formally, the expansion set of concepts \textit{C\textsubscript{p}} is obtained as in equation 3: 
\begin{equation}
\begin{multlined}
C_p =\! \bigcup_{c \in C_s,c' \in C} \{(c',w') : \exists r(s,f)\!=\!c\!\Rightarrow\!{c'}\} \\
subject\ to:\ |c'|=|Y|, s>=\epsilon, f>=\upsilon
\end{multlined}
\end{equation}
Note that we add all the concepts that are implied by \textit{c} where this implication meets the support and confidence thresholds ($\epsilon$, $\upsilon$) respectively. The weight of $c'$ is denoted by $w'$. Two weighting mechanisms can be utilized here: 1) inheritance; where $c'$ will has the same weight as its antecedent $c$, and 2) proportional; where $c'$ will have prorated weight $w'=f*w$ based on the confidence $f$ of $c\!\Rightarrow\!{c'}$.


Finally, all the concepts from search and expansion phases are merged to construct the concept vector \textit{C\textsubscript{t}} of term(s) \textit{t}. We use the disjoint union of $C_s$ and $C_p$ to keep track of all the weights assigned to each concept as in equation 4:
\begin{equation}
C_t = C_s \sqcup C_p
\end{equation}

\subsection{Concept Weighting}
Any concept $c\!\in\!C_t$ should have appeared in $C_s$ at most once, however $c$ might have appeared in $C_p$ multiple times with different weights. Suppose that $\{w_1,...,w_n\}$ denotes all the weights $c$ has in $C_t$ where $n<|C_p|+1$, then we can calculate the final weight $w$ by aggregating all the weights as in equation 5:
\begin{equation}
w = \sum_{i=1}^{n} w_i
\end{equation}

Or we can take the maximum weight as in equation 6:
\begin{equation}
w = \max_i w_i \qquad 1\!\le\!i\!\le\!n
\end{equation}

The rationale behind weight aggregation (equation 5) is to ensure that popular concepts which appear repeatedly in the expansion list will have higher weights than those which are less popular. As this scheme might favor popular concepts even if they appear in the tail of $C_s$ and/or $C_p$, we propose selecting only the maximum weight (equation 6) to ensure that top relevant concepts in both $C_s$ and $C_p$ still have the chance to maintain their high weights.

\subsection{Relatedness Scoring}
We apply the cosine similarity measure in order to calculate the relatedness score between a pair of concept vectors $\mathbf{u}$ and $\mathbf{v}$. Because the concept vectors are sparse, we can rewrite the cosine similarity as in \citet{songunsupervised}. Suppose that $\mathbf{u}\!=\!\{(c_{n_1},u_1),\dotsc,\\(c_{n_{|\mathbf{u}|}},u_{|\mathbf{u}|})\!\}$ and $\mathbf{v}\!=\!\{(c_{m_1},v_1),\dotsc,(c_{m_{|\mathbf{v}|}},v_{|\mathbf{v}|})\!\}$, where $u_i$ and $v_j$ are the corresponding weights of concepts $c_{n_i}$ and $c_{m_j}$. $n_i$ and $m_j$ are the indices of these concepts in the concepts set $C$ such that $1\!\le\!n_i,m_j\!\le\!N$. Then the relatedness score can be written as in equation 7:
\begin{equation}
\begin{multlined}
Rel_{cos}(\mathbf{u},\mathbf{v}) = \frac{\sum_{i=1}^{|\mathbf{u}|}\sum_{j=1}^{|\mathbf{v}|}\mathbbm{1}(n_i\mathord{=}m_j) u_iv_j}{\sqrt{\sum_{i=1}^{|\mathbf{u}|}u_i^2} \sqrt{\sum_{j=1}^{|\mathbf{v}|}v_j^2}}
\end{multlined}
\end{equation}

where $\mathbbm{1}$ is the indicator function which returns 1 if $n_i\mathord{=}m_j$ and 0 otherwise.

\begin{table*}[]
	\centering
	\normalfont
	\caption[Evaluating Semantic Similarity using Various Representation Models]{Evaluating semantic similarity between two highly similar text snippets using various representation models. The BoW is the least successful, while concept space and distributed representation models work relatively better\footnotemark.}
	\begin{tabular}{|c|m{1.7cm}|m{5cm}|m{5cm}|l|}
		\cline{3-4}
		\multicolumn{1}{c}{}& \multicolumn{1}{c|}{}&  \multicolumn{1}{c|}{\textbf{Snippet\#1}}& \multicolumn{1}{c|}{\textbf{Snippet\#2}}& \multicolumn{1}{c}{} \\ \hline
		\textbf{No} & \textbf{Model} & \multicolumn{1}{c|}{\includegraphics[width=6cm,keepaspectratio]{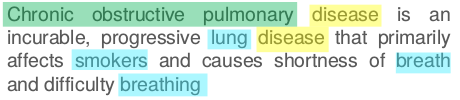}} & \multicolumn{1}{c|}{\includegraphics[width=6cm,keepaspectratio]{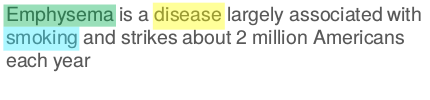}} & \textbf{Similarity} \\ \hline
		A. & BoW & \textcolor{gray}{chronic obstruct pulmonari} \colorbox{yellow}{diseas} \textcolor{gray}{incur progress lung primarili affect smoker caus short breath difficulti} & \textcolor{gray}{emphysema} \colorbox{yellow}{diseas} \textcolor{gray}{larg associ smoke strike million american year} & \multicolumn{1}{c|}{0.09} \\ \hline 
		&&&& \\
		B. & Word2Vec & \multicolumn{1}{c|}{\includegraphics[width=5cm,keepaspectratio]{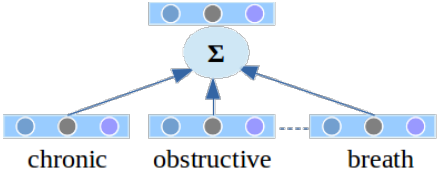}} & \multicolumn{1}{c|}{\includegraphics[width=5cm,keepaspectratio]{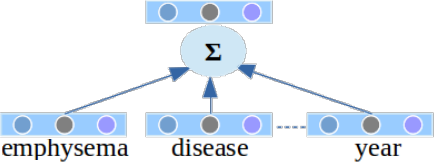}} & \multicolumn{1}{c|}{0.88} \\ \hline
		&&\multicolumn{2}{c|}{}& \\
		C. & MSA & \multicolumn{2}{c|}{$ $\includegraphics[width=10.5cm,keepaspectratio]{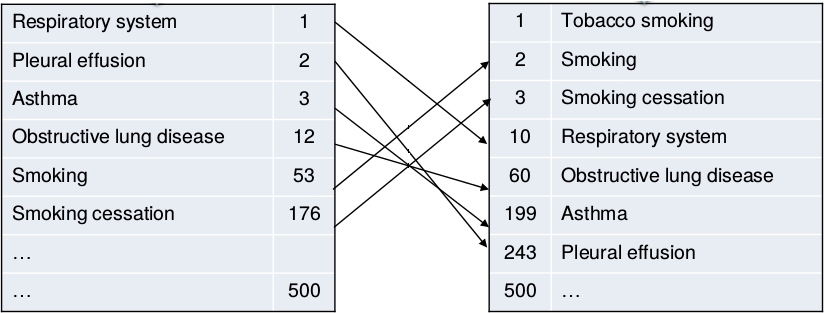}} & \multicolumn{1}{c|}{0.81} \\ \hline
	\end{tabular}
	\label{example}
\end{table*}

Similar to \citet{hassan2011semantic}, we include a normalization factor \begin{math}\lambda\end{math} as the cosine measure gives low scores for highly related terms due to the sparsity of their concept vectors. Other approaches for dealing with vector sparsity worth exploring in the future \cite{songunsupervised}. Using \begin{math}\lambda\end{math}, the final relatedness score will be adjusted as in equation 8:
\begin{equation}
Rel(t_1,t_2) = \begin{cases}
1 & Rel_{cos}(t_1,t_2)\geq\lambda \\
\frac{Rel_{cos}(t_1,t_2)}{\lambda} & Rel_{cos}(t_1,t_2)<\lambda
\end{cases}
\end{equation}

\footnotetext[4]{\emph{Emphysema} is another name to \emph{Chronic obstructive pulmonary disease.}}
\section{A Qualitative Example}
Explicit concept space representations are expressive, easy to interpret, and easy to manipulate and interact with. These characteristics make such models more appealing than other vector-based distributional models. We demonstrate the expressiveness of {\it MSA} by comparing its representation against the traditional BoW and the Word2Vec models for measuring the semantic similarity between two semantically similar text snippets as shown in Table \ref{example}. The color codings in the first row reflect the structures which are identical or semantically similar/related (e.g., {\it disease} in both snippets, and {\it smoking} \& {\it breathing} are related). 

As we can notice in row-A, the BoW model fails to capture these similarities even after stemming and normalizing the two snippets, and gives very low similarity score of 0.09. 

The Word2Vec model in row-B generates the vector representation of each snippet by averaging the embeddings of individual words. Similarity is then measured by comparing the average vectors using the cosine score. As we can notice, though the Word2Vec model gives high score of 0.88\footnote{Using 500 dimension vectors from Word2Vec trained on Wikipedia dump of August 2016.} reflecting the high similarity between the two snippets, it lacks meaning because words are represented as a bunch of real-valued numbers which we cannot interpret. 

{\it MSA} in row-C, overcomes the expressiveness limitation of Word2Vec by generating concept-based representations of the two texts that capture their main topics. For example, concepts like \emph{Asthma}, and \emph{Pleural effusion} capture the lung disease, while concepts like \emph{Smoking}, \emph{Respiratory system}, and \emph{Tobacco smoking} capture the relation between the disease and smoking/breathing. {\it MSA} besides this expressive superiority, gives a similarity score of 0.81\footnote{Using BoC of size 500 concepts.} between the two snippets reflecting their high semantic similarity.

\section{Experiments and Results}
We evaluate the efficacy of {\it MSA} concept vectors on two text analysis tasks: 1) measuring lexical semantic relatedness between pairs of words, and 2) evaluating short text similarity between pairs of short text snippets.

\subsection{Lexical Semantic Relatedness}
\subsubsection{Datasets\protect\footnote{https://github.com/mfaruqui/eval-word-vectors/tree/master/data/word-sim\label{mfaruqui}}}
We evaluate \textit{MSA}'s performance on benchmark datasets for measuring lexical semantic relatedness. Each dataset is a collection of word pairs along with human judged similarity/relatedness score for each pair.

\textbf{\textit{RG}:} A similarity dataset created by \citet{Rubenstein:1965}. It contains 65 noun pairs with similarity judgments 
ranging from 0 (very unrelated) to 4 (very related). 

\textbf{\textit{MC}:} A similarity dataset created by \citet{miller1991contextual}. It contains 30 noun pairs taken from \textit{RG} dataset. 

\textbf{\textit{WS}:} A relatedness dataset of 353 word pairs created by \citet{finkelstein2001placing}. Relatedness scores for each pair 
range from 0 (totally unrelated) to 10 (very related). 

\textbf{\textit{WSS \& WSR}:} \citet{agirre2009study} manually split \textit{WS} dataset into two subsets to separate between similar words (\textit{WSS} of 203 pairs), and related words (\textit{WSR} of 252 pairs). 

\textbf{\textit{MEN}\footnote{http://clic.cimec.unitn.it/~elia.bruni/MEN.html}:} A relatedness dataset created by \citet{bruni2014multimodal}. We use the test subset of this dataset which contains 1000 pairs. Relatedness scores range from 0 to 50. 

\subsubsection{Experimental Setup}
We followed experimental setup similar to \citet{baroni2014don}. Basically, we 
use one of the datasets as a development set for tuning \textit{MSA}'s parameters and then use the tuned parameters to evaluate \textit{MSA}'s performance on the other datasets. Some parameters are fixed with all datasets; namely we set $|Y|\!=\!1$, $\sigma\!=\!1$, and $\upsilon\!=\!0.0$.

We built the search index using a \textit{Wikipedia} dump of August 2016\footnote{http://dumps.wikimedia.org/backup-index.html\label{wiki2016}}. The total uncompressed XML dump size was about 52GB representing about 7 million articles. We extracted the articles using a modified version of Wikipedia Extractor \footnote{http://medialab.di.unipi.it/wiki/Wikipedia\textunderscore Extractor}. Our version \footnote{https://github.com/walid-shalaby/wikiextractor} 
extracts articles plain text discarding images and tables. We also discard \textit{References} and \textit{External links} sections (if any). We pruned both articles not under the main namespace and pruned all redirect pages as well. Eventually, our index contained about 4.8 million documents in total.

\subsubsection{Evaluation}
We report the results by measuring the correlation between \textit{MSA}'s computed relatedness scores and the gold standard provided by human judgments. As in prior studies, we report both Pearson correlation (\textit{r}) \cite{RefWorks:118} and Spearman rank-order correlation (\textit{$\rho$}) \cite{zwillinger1999crc}. 
We compare our results with those obtained from three types of semantic representation models. First, statistical co-occurrence models like \textit{LSA} \cite{landauer1997well}, \textit{CW and BOW} \cite{agirre2009study}, and \textit{ADW} \cite{pilehvar2015senses}. Second, neural network models like Collobert and Weston (\textit{CW}) vectors \cite{collobert2008unified}, \textit{Word2Vec} \cite{baroni2014don}, and \textit{GloVe} \cite{pennington2014glove}. Third, explicit semantics models like \textit{ESA} \cite{gabrilovich2007computing}, \textit{SSA} \cite{hassan2011semantic}, and \textit{NASARI} \cite{camacho2015nasari}.

\begin{table}
	\centering
	\caption{\textit{MSA}'s Pearson (\textit{r}) scores on benchmark datasets vs. other techniques. ($\star$) from \cite{hassan2011semantic}, ($\triangleright$) from \cite{baroni2014don}, ($\diamond$) from \cite{camacho2015nasari}.}
	\label{pearson_all}
	\begin{tabular}{l|lllll} \specialrule{.1em}{.1em}{0.1em}
		&\textit{MC}&\textit{RG}&\textit{WSS}&\textit{WSR}&\textit{WS} \\ \hline
		\textit{LSA}\textsuperscript{$\star$}&0.73&0.64&\multicolumn{1}{c}{--}&\multicolumn{1}{c}{--}&0.56  \\
		\textit{ESA}\textsuperscript{$\diamond$}&0.74&0.72&0.45&\multicolumn{1}{c}{--}&0.49\textsuperscript{$\star$}  \\
		\textit{SSA\textsubscript{s}}\textsuperscript{$\star$}&0.87&0.85&\multicolumn{1}{c}{--}&\multicolumn{1}{c}{--}&0.62  \\
		\textit{SSA\textsubscript{c}}\textsuperscript{$\star$}&0.88&0.86&\multicolumn{1}{c}{--}&\multicolumn{1}{c}{--}&0.59  \\
		\textit{ADW\textsuperscript{$\diamond$}}&0.79&\textbf{0.91}&0.72&\multicolumn{1}{c}{--}&\multicolumn{1}{c}{--}  \\
		\textit{NASARI\textsuperscript{$\diamond$}}&\textbf{0.91}&\textbf{0.91}&0.74&\multicolumn{1}{c}{--}&\multicolumn{1}{c}{--}  \\
		\textit{Word2Vec\textsubscript}\textsuperscript{$\triangleright$}&0.82&0.84&0.76&0.65&0.68 \\ \hline
		\textit{MSA}&\textbf{0.91}&0.87&\textbf{0.77}&\textbf{0.66}&\textbf{0.69}  \\ 
		\specialrule{.1em}{.1em}{0.1em}			
	\end{tabular}
\end{table}

\begin{table}
	\centering	
	\caption{\textit{MSA}'s Spearman (\textit{$\rho$}) scores on benchmark datasets vs. other techniques. ($\star$) from \cite{hassan2011semantic}, ($\ddagger$) from \cite{pilehvar2015senses,hassan2011semantic}, ($\Upsilon$) from \cite{agirre2009study}, ($\S$) from \cite{camacho2015nasari}, ($\diamond$) from \cite{pilehvar2015senses}, ($\Psi$) from \cite{pennington2014glove}, ($\triangleright$) from \cite{baroni2014don}}	
	\label{spearman_all}
	\begin{tabular}{l|lllll} \specialrule{.1em}{.1em}{0.1em}
		&\textit{MC}&\textit{RG}&\textit{WSS}&\textit{WSR}&\textit{WS} \\ \hline
		\textit{LSA}\textsuperscript{$\star$}&0.66&0.61&\multicolumn{1}{c}{--}&\multicolumn{1}{c}{--}&0.58  \\
		\textit{ESA}\textsuperscript{$\ddagger$}&0.70&0.75&0.53&\multicolumn{1}{c}{--}&0.75  \\
		\textit{SSA\textsubscript{s}}\textsuperscript{$\star$}&0.81&0.83&\multicolumn{1}{c}{--}&\multicolumn{1}{c}{--}&0.63  \\
		\textit{SSA\textsubscript{c}}\textsuperscript{$\star$}&0.84&0.83&\multicolumn{1}{c}{--}&\multicolumn{1}{c}{--}&0.60  \\
		\textit{CW\textsuperscript{$\Upsilon$}}&\multicolumn{1}{c}{--}&0.89&{\bf 0.77}&0.46&0.60  \\ 
		\textit{BOW\textsuperscript{$\Upsilon$}}&\multicolumn{1}{c}{--}&0.81&0.70&0.62&0.65  \\ 
		\textit{NASARI\textsuperscript{$\S$}}&0.80&0.78&0.73&\multicolumn{1}{c}{--}&\multicolumn{1}{c}{--}  \\
		\textit{ADW\textsuperscript{$\diamond$}}&{\bf 0.90}\footnotemark&\textbf{0.92}&0.75&\multicolumn{1}{c}{--}&\multicolumn{1}{c}{--}  \\
		\textit{GloVe\textsuperscript{$\Psi$}}&0.84&0.83&\multicolumn{1}{c}{--}&\multicolumn{1}{c}{--}&{\bf 0.76}  \\ 
		\textit{Word2Vec}\textsuperscript{$\triangleright$}&0.82\footnotemark&0.84&0.76&0.64&0.71 \\ \hline
		\textit{MSA}&0.87&0.86&{\bf 0.77}&{\bf 0.71}&0.73  \\ 
		\specialrule{.1em}{.1em}{0.1em}
	\end{tabular}
\end{table}

\subsubsection{Results}
We report \textit{MSA}'s correlation scores compared to other models in Tables \ref{pearson_all}, \ref{spearman_all}, and \ref{men}. Some models do not report their correlation scores on all datasets, so we leave them blank. \textit{MSA} (last row) represents scores obtained by using \textit{WS} as a development set for tuning \textit{MSA}'s parameters and evaluating performance on the other datasets using the tuned parameters. The parameter values obtained by tuning on \textit{WS} were $L\!=\!5k$, $O\!=\!1$, $M\!=\!800$, $\tau\!=\!2, 3$ for \textit{C\textsubscript{s}}, \textit{C\textsubscript{p}} respectively, and finally $\epsilon\!=\!1$.

Table \ref{pearson_all} shows \textit{MSA}'s Pearson correlation (\textit{r}) on five benchmark datasets compared to prior work. For \textit{Word2Vec}, we obtained \citet{baroni2014don} predict vectors\textsuperscript{\ref{semvecs}} and used them to calculate Pearson correlation scores. It is clear that, in absolute numbers, \textit{MSA} consistently gives the highest correlation scores on all datasets compared to the other methods except on \textit{RG} where \textit{NASARI} and \textit{ADW} \cite{camacho2015nasari} performed better.

\footnotetext[12]{Pairwise similarity scores obtained from \citet{pilehvar2015senses}} 
\footnotetext[13]{Using http://clic.cimec.unitn.it/composes/semantic-vectors.html\label{semvecs}}


Table \ref{spearman_all} shows \textit{MSA}'s Spearman correlation scores compared to prior models on same datasets as in Table \ref{pearson_all}. As we can see, \textit{MSA} gives highest scores on \textit{WSS} and \textit{WSR} datasets. It comes second on \textit{MC}, third on \textit{RG} and \textit{WS}. We can notice that {\it MSA}'s concept enrichments participated in performance gains compared to other explicit concept space models such as \textit{ESA} and \textit{SSA}. In addition, \textit{MSA} consistently outperformed the popular \textit{Word2Vec} model on all datasets.


\begin{table}
	\centering
	\caption{\textit{MSA}'s Spearman (\textit{$\rho$}) scores on \textit{MEN} dataset vs. other techniques. ($\star$) from \cite{hill2014not}, ($\triangleright$) from \cite{baroni2014don}}
	\label{men}
	\begin{tabular}{l|l} \specialrule{.1em}{.05em}{0.05em}
		&\textit{MEN} \\ \hline			
		\textit{Skip-gram\textsuperscript{$\star$}}&0.44  \\ 
		\textit{CW}\textsuperscript{$\star$}&0.60  \\  
		\textit{GloVe}\textsuperscript{$\star$}&0.71  \\  
		\textit{Word2Vec}\textsuperscript{$\triangleright$}&{\bf 0.79}  \\  \hline
		\textit{MSA}&0.75  \\  
		\specialrule{.1em}{.1em}{0.1em}
	\end{tabular}
\end{table}

Table \ref{men} shows \textit{MSA}'s Spearman correlation score vs. other models on {\it MEN} dataset. As we can see, \textit{MSA} comes second after \textit{Word2Vec} giving higher correlation than \textit{Skip-gram}, \textit{CW}, and \textit{GloVe}. Results on this dataset prove that \textit{MSA} is a very advantageous method for evaluating lexical semantic relatedness compared to the popular deep learning models. On another hand, \textit{MSA}'s Pearson correlation score on \textit{MEN} dataset was 0.73.

We can notice from the results in Tables \ref{pearson_all} and Table \ref{spearman_all} that measuring semantic relatedness is more difficult than measuring semantic similarity. This is clear from the drop in correlation scores of the relatedness only dataset (\textit{WSR}) compared to the similarity only datasets (\textit{MC, RG, WSS}). This pattern is common among \textit{MSA} and all prior techniques which report on these datasets.

\subsection{Short Text Similarity}
\subsubsection{Dataset}
We evaluate {\it MSA}'s performance for scoring pairwise short text similarity on Lee50\footnote{http://faculty.sites.uci.edu/mdlee/similarity-data/} dataset \cite{lee2005empirical}. The dataset contains 50 short documents collected from the Australian Broadcasting Corporation's news mail service. On average each document has about 81 words. Human annotators were asked to score the semantic similarity of each document to all other documents in the collection. As in previous work, we averaged all human similarity ratings for the same document pair to obtain single score for each pair. This resulted in 1225 unique scored pairs.
\subsubsection{Experimental Setup}
We followed experimental setup similar to \citet{song2014dataless,songunsupervised} for fair comparison. Specifically we created a \textit{Wikipedia} index using  August 2016 dump\textsuperscript{\ref{wiki2016}}. We indexed all articles whose length is at least 500 ($L\!=\!500$) and has at least 30 outgoing links ($O\!=\!30$) using the code base of dataless hierarchical text classification\footnote{http://cogcomp.cs.illinois.edu/page/software\_view/DatalessHC\label{DatalessHierarchicalTextClassification}}. As previously we set $|Y|\!=\!1$, $\upsilon\!=\!0.0$. We also set $\sigma\!=\!5$ and relaxed $\tau$. 

\subsubsection{Evaluation}
We report the both Pearson ($r$) and Spearman ($\rho$) correlations between {\it MSA}'s similarity scores and human judgments using a concept vector of size 500 ($M\!=\!500$) as in \citet{songunsupervised}. We compare our results to {\it ESA}, {\it LSA}, and {\it SSA}.

\begin{table}
	\centering
	\caption{{\it MSA}'s Spearman (\textit{$\rho$}) and Pearson ($r$) scores on Lee50 dataset vs. other techniques. ($\star$) from \cite{hassan2011semantic}}
	\label{textsim}
	\begin{tabular}{l|lll|l} \specialrule{.1em}{.1em}{0.1em}
		&\textit{LSA}\textsuperscript{$\star$}&\textit{SSA}\textsuperscript{$\star$}&\textit{ESA}\textsuperscript{\ref{DatalessHierarchicalTextClassification}}&\textit{MSA} \\ \hline
		\textit{$Spearman (\rho$)}&0.46&0.49&0.61&{\bf 0.62}  \\ \hline
		\textit{$Pearson (r)$}&0.70&0.68&0.73&{\bf 0.75}  \\
		\specialrule{.1em}{.1em}{0.1em}
	\end{tabular}
\end{table}

\begin{figure*}[htb]
	\centering
	\begin{tabular}{@{}cc@{}}
		\includegraphics[width=.437\textwidth]{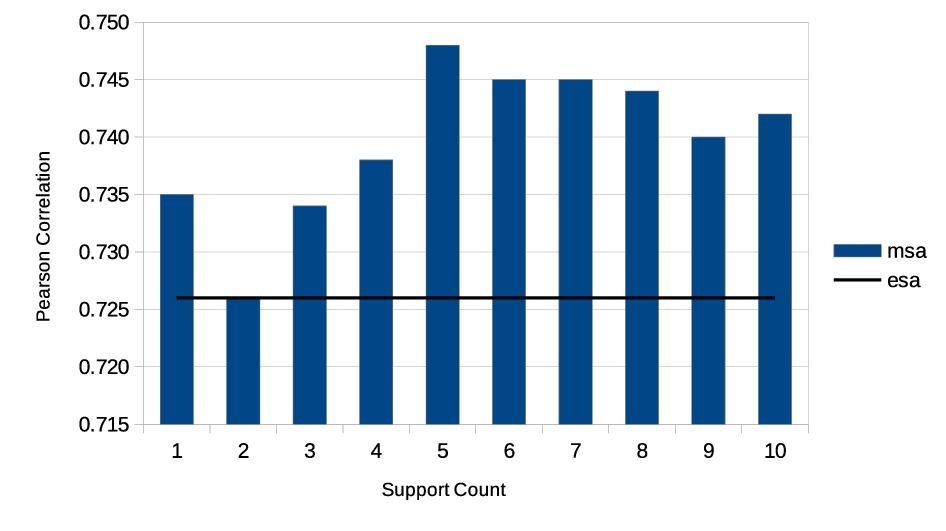} &
		\includegraphics[width=.437\textwidth]{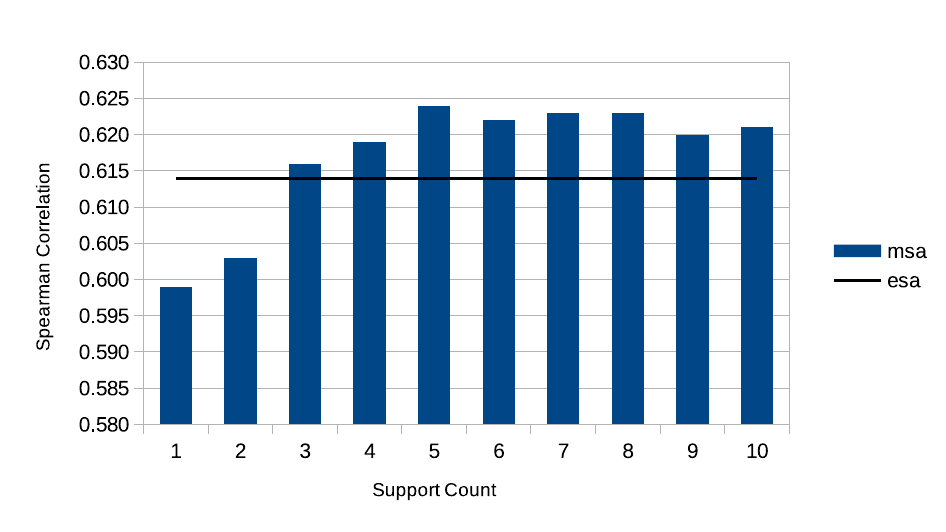}  \\ 
		(a) & (b) \\
		\includegraphics[width=.437\textwidth]{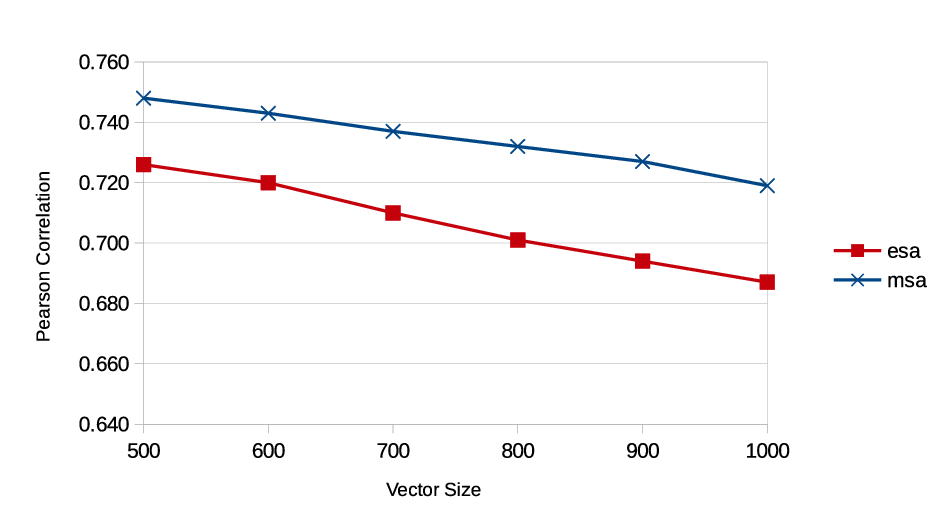} &
		\includegraphics[width=.437\textwidth]{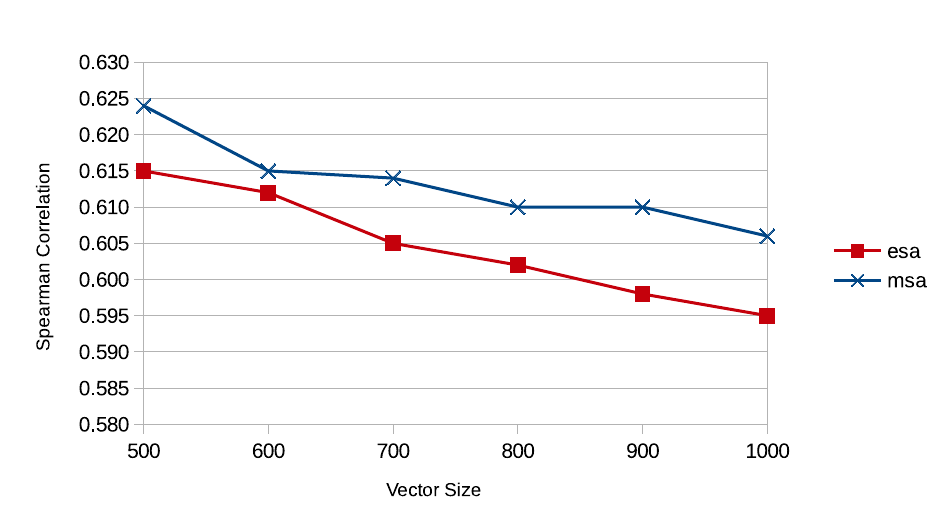}   \\
		(c) & (d)
	\end{tabular}
	\caption{{\it MSA}'s correlation scores on Lee50 dataset. (a) Pearson ($r$) correlation when varying support count, (b) Spearman ($\rho$) correlation when varying support count, (c) Pearson ($r$) correlation when varying vector size, and (d) Spearman ($\rho$) correlation when varying vector size.}
	\label{analysis}
\end{figure*}

\subsubsection{Results}
As we can see in Table \ref{textsim}, {\it MSA} outperforms prior models on Lee50 dataset. {\it ESA} comes second after {\it MSA} which shows the potential of concept space augmentation using implicitly associated concepts discovered through {\it MSA}'s concept-concept association learning.

We performed another experiment in order to assess the impact of parameter tuning on the reported results in Table \ref{textsim}. Figures \ref{analysis}-a and \ref{analysis}-b show {\it MSA}'s $r$ and $\rho$ correlation scores when varying the support count parameter ($\sigma$) within the 1 to 10 range in steps of 1. As we can see {\it MSA}'s $r$ correlation scores are consistently higher than {\it ESA}. In addition, $\rho$ correlation scores of {\it MSA} are higher than {\it ESA} score for all $\sigma$ values between 3 and 10. Figures \ref{analysis}-c and \ref{analysis}-d show {\it MSA}'s $r$ and $\rho$ correlation scores compared to {\it ESA} when varying the vector size parameter ($M$) within the 500 to 1000 range in steps of 100. As we can see both $r$ and $\rho$ correlation scores of {\it MSA} are consistently higher than {\it ESA}. 

Two other observations we can notice from Figures \ref{analysis}-c and \ref{analysis}-d. First, both $r$ and $\rho$ scores of {\it MSA} and {\it ESA} tend to decrease as we increase the vector size. This is because more irrelevant concepts are added to the concept vector causing divergence from a "better" to a "worse" representation of the given document. Second, and more importantly, {\it MSA}'s $r$ and $\rho$ correlation scores are higher than {\it SSA} and {\it LSA} over all values of $\sigma$ and $M$. This experiment reflects that {\it MSA}'s semantic representations are largely robust against parameters tuning.



\begin{table*}
	\centering
	\caption{Steiger's Z significance test on the differences between Spearman correlations ($\rho$) using 1-tailed test and 0.05 statistical significance. ($\triangleright$) using \cite{baroni2014don} predict vectors, ($\star$) using \cite{camacho2015nasari} pairwise similarity scores, ($\diamond$) using \cite{pilehvar2015senses} pairwise similarity scores.}
	\label{sig}
	\begin{tabular}{l|cc|cc|cc|cc|cc|cc} \specialrule{.1em}{.05em}{0.05em}
		&\multicolumn{2}{c}{\textit{MC}}&\multicolumn{2}{c}{\textit{RG}}&\multicolumn{2}{c}{\textit{WSS}}&\multicolumn{2}{c}{\textit{WSR}}&\multicolumn{2}{c}{\textit{WS}}&\multicolumn{2}{c}{\textit{MEN}} \\ \hline
		&$\rho$&\textit{p}-value&$\rho$&\textit{p}-value&$\rho$&\textit{p}-value&$\rho$&\textit{p}-value&$\rho$&\textit{p}-value&$\rho$&\textit{p}-value \\ \hline
		&\multicolumn{12}{c}{\textit{MSA\textsubscript{t}}} \\ \hline		
		\textit{Word2Vec\textsubscript{t}}\textsuperscript{$\triangleright$}&0.84&0.168&0.78&0.297&0.79&0.357&0.70&{\bf 0.019}&0.72&0.218&0.78&{\bf 0.001} \\ 
		\textit{NASARI}\textsuperscript{$\star$}&0.73&0.138&0.77&{\bf 0.030}&0.70&0.109&--&--&--&--&--&-- \\ 
		\textit{ADW}\textsuperscript{$\diamond$}&0.78&0.258&0.78&{\bf 0.019}&0.67&0.271&--&--&--&--&--&-- \\ \specialrule{.1em}{.1em}{.1em}
		&\multicolumn{12}{c}{\textit{ADW}\textsuperscript{$\diamond$}} \\ \hline		
		\textit{Word2Vec\textsubscript{t}\textsuperscript{$\triangleright$}}&0.80&0.058&0.81&{\bf 0.003}&0.68&0.5&--&--&--&--&--&-- \\
		\textit{NASARI}\textsuperscript{$\star$}&0.82&{\bf 0.025}&0.80&{\bf 0.0}&0.76&0.256&--&--&--&--&--&-- \\ 
		\specialrule{.1em}{.1em}{.1em}
		&\multicolumn{12}{c}{\textit{Word2Vec\textsubscript{t}}\textsuperscript{$\triangleright$}} \\ \hline		
		\textit{NASARI}\textsuperscript{$\star$}&0.75&0.387&0.71&0.105&0.66&0.192&--&--&--&--&--&-- \\  \specialrule{.1em}{.1em}{.1em}
	\end{tabular}
\end{table*}

\section{A Study on Statistical Significance}
Through the results section, we kept away from naming state-of-the-art method. That was due two facts. First, the differences between reported correlation scores were small. Second, the size of the datasets was not that large to accommodate for such small differences. These two facts raise a question about the statistical significance of improvement reported by some method A compared to another well performing method B. 

We hypothesize that the best method is not necessarily the one that gives the highest correlation score. In other words, being state-of-the-art does not require giving the highest correlation, rather giving a relatively high score that makes any other higher score statistically insignificant.

To test our hypothesis, we decided to perform statistical significance tests on the top reported correlations. Initially we targeted \textit{Word2Vec}, \textit{GloVe}, \textit{ADW}, and \textit{NASARI} besides \textit{MSA}. We contacted several authors and some of them thankfully provided us with pairwise relatedness scores on the corresponding benchmark datasets. 

To measure statistical significance, we performed Steiger's Z significance test \cite{steiger1980tests}. This test evaluates whether the difference between two dependent correlations obtained from the same sample is statistically significant or not, i.e., whether the two correlations are statistically equivalent. 

Steiger's Z test requires to calculate the correlation between the two correlations. We applied the tests on Spearman correlations ($\rho$) as it is more commonly used than Pearson (\textit{r}) correlation. We conducted the tests using correlation scores of \textit{MSA}'s tuned model on \textit{WS} dataset, \textit{Word2Vec}, \textit{ADW}, and \textit{NSASRI}.

Table \ref{sig}, shows the results using 1-tailed test with significance level 0.05. For each dataset, we report method-method Spearman correlation ($\rho$) calculated using reported scores in Table \ref{spearman_all} and Table \ref{men}. We report \textit{p}-value of the test as well. 

On \textit{MC} dataset, the difference between \textit{MSA} score and all other methods was statistically insignificant.  Only \textit{ADW} score was statistically significant compared to \textit{NSASARI}. This implies that {\it MSA} can be considered statistically a top performer on \textit{MC} dataset.

On \textit{RG} dataset, \textit{MSA} gave significant improvement over \textit{NASARI}. \textit{ADW} score was significantly better than \textit{Word2Vec}, \textit{NASARI}, and \textit{MSA}. Overall, \textit{ADW} can be considered the best on \textit{RG} dataset followed by \textit{MSA} and \textit{Word2Vec} (their $\rho$ scores are 0.92, 0.86, and 0.84 respectively). 

On \textit{WSS}, though \textit{MSA} achieved the highest score ($\rho$=0.77), no significant improvement was proved. Therefore, the differences between the four methods can be considered statistically insignificant.

On \textit{WSR}, \textit{WS}, and \textit{MEN} datasets, we could obtain pairwise relatedness scores of \textit{Word2Vec} only. The significance test results indicated that, the improvement of \textit{MSA} over \textit{Word2Vec} on \textit{WS} was statistically insignificant (their $\rho$ scores are 0.77 and 0.76 respectively). On the other hand, \textit{MSA} was statistically better than \textit{Word2Vec} on \textit{WSR} dataset (their $\rho$ scores are 0.71 and 0.64 respectively), while \textit{Word2Vec} was statistically better than \textit{MSA} on \textit{MEN} dataset (their $\rho$ scores are 0.79 and 0.75 respectively).

This comparative study is one of the main contributions of this paper. To our knowledge, this is the first study that addresses evaluating the statistical significance of results across various semantic relatedness methods. Additionally, this study positions \textit{MSA} as a state-of-the-art method for measuring semantic relatedness compared to other explicit concept-based representation methods such as \textit{ESA} and \textit{SSA}. It also shows that \textit{MSA} is very competitive to other neural-based representations such as \textit{Word2Vec} and \textit{GloVe}.

\section{Conclusion}
In this paper, we presented \textit{MSA}; a novel approach for semantic analysis which employs data mining techniques to create conceptual vector representations of text. \textit{MSA} is motivated by inability of prior concept space models to capture implicit relations between concepts. To this end, \textit{MSA} mines for implicit concept-concept associations through \textit{Wikipedia}'s \textit{"See also"} link graph. 

Intuitively, \textit{"See also"} links represent related concepts that might complement the conceptual knowledge about a given concept. Furthermore, it is common in most online encyclopedic portals to have a "{\it See also}" or "{\it Related Entries}" sections opening the door for more conceptual knowledge augmentation using these resources in the future.

Through empirical results, we demonstrated \textit{MSA}'s effectiveness to measure lexical semantic relatedness on benchmark datasets. In absolute numbers, \textit{MSA} could consistently produce higher Pearson correlation scores than other explicit concept space models such as \textit{ESA}, \textit{SSA} on all data  sets. Additionally, \textit{MSA} could produce higher scores than \textit{ADW} and \textit{NASARI} on four out of five datasets. On another hand, \textit{MSA} scores were higher than predictive models built using neural networks such as \textit{Word2Vec}. 

Regarding Spearman correlation, \textit{MSA} produced the highest scores on two datasets (\textit{WSS} and {\textit{WSR}). Results on other datasets were very competitive in absolute numbers. 
Additionally, \textit{MSA} gave higher correlation on \textit{MEN} dataset than neural-based representations including \textit{Skip-gram}, \textit{CW}, and \textit{GloVe}.
	
The results show \textit{MSA} competitiveness compared to state-of-the-art methods. More importantly, our method produced significantly higher correlation scores than previous explicit semantics methods (\textit{ESA} and \textit{SSA}). The good performance demonstrates the potential of \textit{MSA} for augmenting the explicit concept space by other semantically related concepts which contribute to understanding the given text.
	
In this paper, we introduced the first comparative study which evaluates the statistical significance of results from across top performing semantic relatedness methods. We used Steiger's Z significance test to evaluate whether reported correlations from two different methods are statistically equivalent even if they are numerically different. We believe this study will help the research community to better evaluate and position state-of-the-art techniques at different application areas. The study proved that, statistically, \textit{MSA} results are either better than or equivalent to state-of-the-art methods on all datasets except \textit{RG} where \textit{ADW} was better, and \textit{MEN} where \textit{Word2Vec} was better.
	
\textit{MSA} is a general purpose semantic representation approach which builds explicit conceptual representations of textual data. We argue that the expressiveness and interpretability of \textit{MSA} representation make it easier for humans to manipulate and interact with. These two advantages favors \textit{MSA} over the popular {\it Word2Vec} representation. Therefore, \textit{MSA} could be leveraged in many text understanding applications which require interactivity and visualization of the underlying representation such as interactive semantic search, concept tracking, technology landscape analysis, and others \cite{shalabyinnomsaflairs2016}. In the future, we also plan to evaluate {\it MSA} on other tasks like textual entailment and word sense disambiguation. 

\textit{MSA} is an efficient technique because it employs an inverted search index to retrieve semantically related concepts to a given text. Additionally, mining for concept(s)-concept(s) association rules is done offline making it scalable to huge amounts of data. 



\bibliographystyle{unsrtnat}
\bibliography{msa}
%



\end{document}